\documentclass[siggraph]{acmart}
\AtBeginDocument{%
  }

\setcopyright{acmlicensed}
\copyrightyear{2025}
\acmYear{2025}
\begin{document}

\title{Leveraging AI to Accelerate Medical Data Cleaning: A Comparative Study of AI-Assisted vs. Traditional Methods}

\author{Matthew Purri, PhD}
\affiliation{%
  \institution{Octozi}
  \city{New York City}
  \state{New York}
  \country{USA}
}
\author{Amit Patel}
\affiliation{%
  \institution{Octozi}
  \city{New York City}
  \state{New York}
  \country{USA}
}
\author{Erik Deurell, MD}
\affiliation{%
  \institution{Octozi}
  \city{New York City}
  \state{New York}
  \country{USA}
}

\renewcommand{\shortauthors}{Purri et al.}

\begin{abstract}
Clinical trial data cleaning represents a critical bottleneck in drug development, with manual review processes struggling to manage exponentially increasing data volumes and complexity. This paper presents Octozi, an artificial intelligence-assisted platform that combines large language models with domain-specific heuristics to transform medical data review. In a controlled experimental study with experienced medical reviewers (n=10), we demonstrate that AI assistance increased data cleaning throughput by 6.03-fold while simultaneously decreasing cleaning errors from 54.67\% to 8.48\% (a 6.44-fold improvement). Crucially, the system reduced false positive queries by 15.48-fold, minimizing unnecessary site burden. Economic analysis of a representative Phase III oncology trial reveals potential cost savings of \$5.1 million, primarily driven by accelerated database lock timelines (5-day reduction saving \$4.4M), improved medical review efficiency (\$420K savings), and reduced query management burden (\$288K savings). These improvements were consistent across reviewers regardless of experience level, suggesting broad applicability. Our findings indicate that AI-assisted approaches can address fundamental inefficiencies in clinical trial operations, potentially accelerating drug development timelines such as database lock by 33\% while maintaining regulatory compliance and significantly reducing operational costs. This work establishes a framework for integrating AI into safety-critical clinical workflows and demonstrates the transformative potential of human-AI collaboration in pharmaceutical clinical trials.

\end{abstract}

\maketitle

\section{Introduction}
The pharmaceutical industry faces an unprecedented data management crisis. Clinical trials continue to generate more data than a decade ago \cite{TAVES2010180}, yet the fundamental processes for ensuring data quality remain largely unchanged since the advent of electronic data capture (EDC). This disconnect between data volume and processing capability creates substantial delays in drug development, with each day of database lock delay costing \$840,000 in lost revenue across an oncology product's lifecycle \cite{smith2024new}. More critically, inadequate data cleaning risks missing crucial safety signals that could impact patient welfare.

Medical data cleaning, the systematic process of reviewing, validating, and reconciling data collected from investigative sites, serves as the foundation for regulatory submissions to authorities including the US Food and Drug Administration (FDA) and European Medicines Agency (EMA). This process extends beyond simple error detection to encompass complex medical judgment, requiring specialized expertise to identify clinically meaningful discrepancies while maintaining regulatory compliance. The stakes are substantial: data quality issues represent one of the most common causes of regulatory findings, potentially delaying or preventing drug approval.

Traditional approaches to medical data cleaning rely heavily on manual processes that have remained fundamentally unchanged for decades. Medical data managers deploy programmatic edit checks to identify basic inconsistencies, while medical monitors conduct labor-intensive reviews of individual patient narratives, laboratory values, and case report forms (CRFs). This manual approach suffers from several critical limitations: inconsistent application of clinical judgment across reviewers, inability to scale with increasing data volumes, and susceptibility to human error during repetitive tasks. Furthermore, the cognitive burden of synthesizing information across multiple data sources often leads to missed relationships and delayed identification of safety signals.

The advent of artificial intelligence, particularly large language models (LLMs) trained on biomedical literature, offers a paradigm shift in clinical data management. These models demonstrate remarkable capability in understanding medical terminology, identifying semantic relationships, and applying consistent logic across vast datasets \cite{parkapplication, tu2024towards}. When combined with domain-specific algorithms and clinical heuristics, AI systems can potentially automate routine aspects of data review while augmenting human expertise for complex medical decisions.

Despite this promise, the integration of AI into clinical workflows faces significant challenges. Regulatory bodies require transparent, auditable processes for safety-critical decisions. Medical reviewers express concerns about "black-box" algorithms making medical judgments. Perhaps most importantly, any AI system must demonstrate not only efficiency gains, but also maintenance or improvement of data quality standards to gain acceptance in this highly regulated environment.
Here, we present Octozi, an AI-assisted medical data cleaning platform designed to address these challenges through a hybrid approach combining machine learning with expert-encoded clinical logic. Our system leverages unified data pipelines to continuously ingest and contextualize diverse clinical trial data, presenting AI-generated insights through an intuitive interface that maintains human oversight while dramatically improving efficiency. We hypothesized that this human-AI collaboration model would improve both the speed and accuracy of medical data review compared to traditional methods.

To rigorously test this hypothesis, we conducted a controlled experimental study with experienced medical reviewers, comparing the performance between traditional spreadsheet-based review and AI-assisted workflows. Our results demonstrate that AI assistance fundamentally can shift the focus of clinical development teams from reactive data cleaning to proactive safety and efficacy monitoring. By dramatically reducing the time spent on routine data reconciliation, AI enables teams to engage in higher value activities: identifying safety signals earlier, optimizing trial protocols in real time, and making more informed decisions about patient care. This transformation not only accelerates database-lock but enables more adaptive, responsive trials that can pivot based on emerging data patterns, ultimately reducing both development risk and time to market for life-saving therapies.

\section{Related Work}

Clinical trial data quality remains a persistent challenge as data volumes and complexity continue to grow exponentially. Modern oncology trials generate millions of data points across dozens of interconnected systems, yet the fundamental processes for ensuring data quality rely on methods developed decades ago for much smaller datasets. This mismatch between data scale and review capabilities creates substantial delays in drug development and risks missing critical safety signals. Various technological approaches have emerged to modernize data review, though each faces significant limitations that prevent comprehensive adoption.

Visualization platforms like JReview and Spotfire enable identification of numerical outliers through scatter plots and heat maps, but face fundamental constraints. These tools cannot process unstructured text data, which comprises the majority of critical data fields such as adverse event narratives and concomitant medication descriptions. Moreover, statistical outliers identified through visualization do not necessarily represent clinically meaningful discrepancies. A laboratory value may be statistically unusual but clinically appropriate given a patient's medical context, while dangerous drug interactions may appear normal in aggregate visualizations.

Recent advances in machine learning have enabled more sophisticated programmatic edit checks that use historical data to calibrate thresholds and reduce false positives. However, these systems remain limited by Boolean logic that cannot extend to unstructured text. They cannot perform the complex medical reasoning required to determine whether an adverse event is consistent with a patient's medical history. The inability to process narrative descriptions means these systems miss critical relationships that are often documented only in free-text fields.

Risk-Based Quality Management (RBQM) approaches, endorsed by FDA and EMA in 2013, use statistical algorithms to identify sites or patterns deviating from expected norms \cite{stoltz2013risk}. While conceptually appealing, these methods require substantial sample sizes to achieve adequate statistical power, rendering them less effective for early trial phases or rare disease studies. Furthermore, like visualization approaches, statistical monitoring cannot process narrative text, and statistical significance does not equate to clinical significance. An RBQM solution, for instance, could not alert a medical reviewer that a concomitant medication was inappropriate for treating a documented adverse event, as this requires understanding the medical relationship between free-text descriptions rather than detecting statistical anomalies.

Our approach with Octozi represents a fundamental departure from these existing paradigms. Rather than adapting general-purpose tools to clinical data, we developed a purpose-built system that combines large language models with deterministic clinical algorithms and external knowledge sources such as FDA drug labels. This hybrid architecture enables several key innovations. First, our LLM-based approach natively processes both structured and unstructured data, identifying discrepancies invisible to traditional methods. For example, recognizing when a narrative description of "severe nausea requiring hospitalization" contradicts a coded severity of "mild." Second, integration of external medical knowledge enables true clinical reasoning rather than pattern matching, identifying contraindications and drug interactions that would appear normal in statistical analyses. Third, unlike statistical methods requiring minimum sample sizes, our system scales seamlessly from single-patient review to population analysis through consistent application of clinical logic. Finally, the platform contextualizes data within the patient's journey, differentiating between expected clinical progressions and true discrepancies recognizing, for example, that platelet drops after the initiation of chemotherapy are expected while the same change without clinical context warrants review. This purpose-built approach addresses the fundamental mismatch between existing tools and medical review requirements, offering a solution that enhances clinical expertise while dramatically improving efficiency and consistency.

\section{Methods}\label{sec2}

\subsection{Study Design and Statistical Considerations}
We designed a within-subjects controlled experiment to evaluate the impact of AI assistance on medical data review performance. This design allowed each participant to serve as their own control, minimizing inter-individual variability and thereby ensuring that observed differences reflected method effects rather than participant differences. Prior to the study, we conducted an a priori power analysis to determine an appropriate sample size for detecting a meaningful improvement in clinical data cleaning throughput. Assuming a conservative large effect size (Cohen’s d = 1.0), a two-sided paired t-test with $\alpha$ = 0.05 would require a minimum of 10 participants to achieve 80\% statistical power. This threshold was selected to detect at least a 50\% improvement in throughput, which we considered a clinically meaningful gain based on early pilot observations.

Upon completion of the study, the observed effect size was substantially larger than anticipated. Based on the actual performance data, the mean improvement in throughput between baseline and AI-augmented methods was 17.1 records per session (standard deviation = 8.16), yielding an observed Cohen’s d of 2.10. A post hoc power analysis using these values confirmed that our sample of 10 participants yielded a power greater than 99.9\% to detect a statistically significant difference in performance ($\alpha$ = 0.05, two-sided). This result confirms that the study was more than adequately powered to evaluate the efficacy of the AI-assisted system.

Study participants were recruited from a diverse pool of clinical research professionals through targeted outreach to pharmaceutical companies, contract research organizations, and academic medical centers. Eligibility criteria included: (1) a minimum two years of experience in medical data review roles, (2) proficiency in adverse event adjudication, and (3) familiarity with oncology medical data cleaning. Pre-study assessments (Appendix B) captured detailed professional backgrounds including medical data cleaning experience, familiarity with specific data management systems, and work experience. 

\begin{table*}[]
    \caption{Summary of clinically meaningful discrepancy categories introduced into the synthetic dataset. These six categories were identified through expert consultation as the most common adverse event-related data quality issues encountered during oncology clinical trial reviews.}
    \label{tab:discrepancy}
    \begin{tabular}{@{}ll@{}}
        \toprule
        & Discrepancy categories \\ \midrule
        1 & Inappropriate concomitant medication to treat an adverse event \\
        2 & Timing of concomitant medication administration and adverse event do not align \\
        3 & Incorrect severity score attached to an adverse event based on the description of the event \\
        4 & Mismatched dosing change \\
        5 & Incorrect causality assessment of adverse event \\
        6 & No supporting data for adverse event \\
        \bottomrule
    \end{tabular}
\end{table*}

\subsection{Synthetic Dataset Construction and Validation}
The experimental datasets were derived from a comprehensive phase III oncology trial database containing electronic data capture (EDC) information across 51 separate case report form datasets for over 150 patients. Given our focus on adverse event discrepancy detection, we strategically selected a subset of 8 CRFs that were directly relevant to adverse event assessment and documentation. This focused approach ensured that participants could concentrate on clinically meaningful data relationships without being overwhelmed by extraneous information.

To create synthetic patients while preserving complete anonymity, we utilized a library-based refinement generation approach. We first constructed comprehensive libraries of clinical elements by extracting all unique adverse events, concomitant medications, ancillary procedures, and medical history entries from the original dataset. These libraries preserved the diversity and complexity of real clinical data while completely dissociating individual data points from their original patient contexts. For each synthetic patient, we randomly sampled entries from these libraries, with sampling frequencies weighted by their occurrence rates in the original data to maintain realistic clinical patterns.

The initial synthetic data underwent rigorous clinical refinement by expert annotators who reviewed each generated patient profile for medical coherence. These experts made targeted modifications to enhance realism, ensuring that temporal relationships, severity progressions, and treatment patterns reflected authentic clinical scenarios. Laboratory values and vital signs were particularly carefully constructed, while initially set to normal ranges, these values were systematically modified to align with documented adverse events. For instance, patients with documented anemia had their hemoglobin values adjusted to clinically appropriate low levels, maintaining internal consistency throughout the dataset.

To simulate realistic data cleaning scenarios, we systematically introduced clinically meaningful discrepancies based on extensive consultation with industry experts. Through structured interviews with three senior clinical scientists and medical monitors, we identified the most common and impactful data quality issues encountered in oncology trials. These interviews revealed six primary categories of discrepancies that represent the majority of clinically significant data quality issues in modern trials (detailed in Table \ref{tab:discrepancy}). Discrepancies were introduced into 10\% of all data points using a stratified randomization approach, ensuring equal distribution across both experimental conditions and discrepancy types.

\subsection{Experimental Protocol}
The study protocol was designed to minimize bias and consisted  of three carefully structured phases. The baseline assessment phase began with participants receiving standardized training on the dataset structure and common discrepancy types. Following a 10-minute briefing period, participants conducted manual reviews using industry-standard spreadsheet tools, with time tracking initiated immediately after the orientation and participant setup of their spreadsheet tool environment. This approach replicated typical medical data review workflows used across the pharmaceutical industry.

The AI-assisted review phase commenced with an interactive tutorial on the Octozi platform features, including live demonstrations of the system's capabilities and hands-on practice with a specially designed training dataset containing known discrepancies. After confirming proficiency with the platform, participants were allotted 30 minutes to review a matched complexity dataset using Octozi's full functionality.

The final phase consisted of standardized assessment using two validated surveys. The inverse NASA Task Load Index (NASA-TLX) was administered to quantify the cognitive workload associated with each review method across six dimensions: mental demand, physical demand, temporal demand, performance, effort, and frustration. This multidimensional assessment was particularly relevant given the cognitive intensity of medical data review and the potential for AI to reduce mental burden. The System Usability Scale (SUS) provided a standardized measure of the Octozi platform's usability, generating a composite score that enabled benchmarking against industry standards for enterprise software. Together, these instruments captured both the cognitive impact of AI assistance and the practical usability of the technology, providing crucial insights into the feasibility of widespread adoption in clinical research settings.

\begin{figure}
    \centering
    \includegraphics[width=1.0\linewidth]{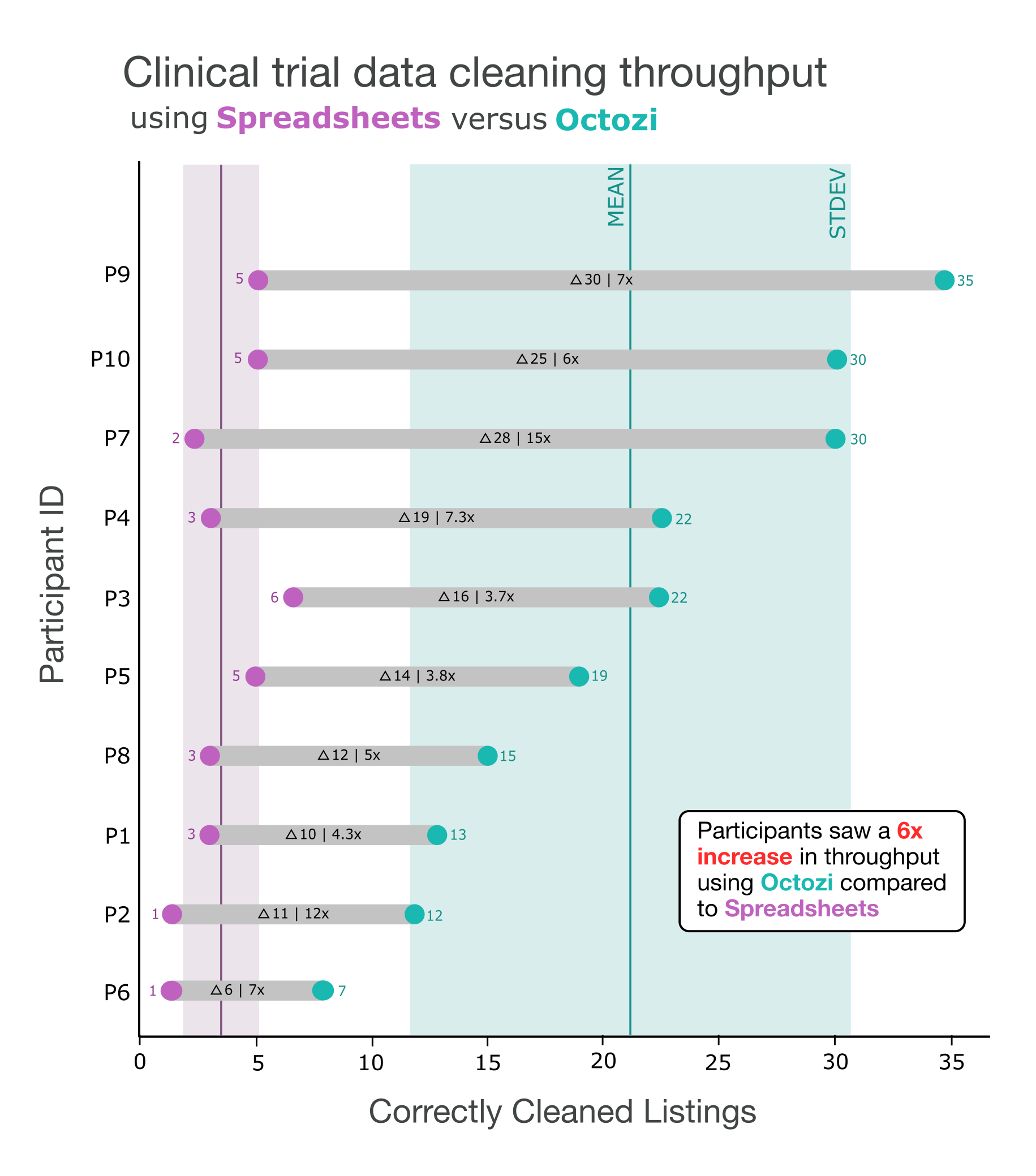}
    \caption{Individual participant throughput in correctly cleaned clinical data listings using traditional spreadsheets versus the AI-assisted Octozi platform. Across all participants, Octozi substantially increased data cleaning throughput, with a mean improvement of 6.03-fold (range: 3.7x to 15x). Participants cleaned a mean of 3.4 data points (range: 1–6) with spreadsheets and a median of 20.5 data points (range: 7–35) using Octozi within the same 30-minute review period. Improvements were observed consistently across all participants, highlighting the effectiveness of AI assistance in medical data review tasks.}
    \label{fig:participant_improvement}
\end{figure}

\subsection{AI-Assisted Medical Data Review Platform}
The Octozi platform represents a comprehensive AI-assisted solution designed to transform medical data review through intelligent data integration, contextual analysis, and automated discrepancy detection. The system architecture comprises four core components that work synergistically to enhance reviewer efficiency and accuracy.

\subsubsection{Data Integration and Harmonization}
The platform ingests and harmonizes data from multiple clinical trial sources, including electronic data capture (EDC) systems, safety databases, central laboratory systems, and patient-reported outcome (PRO) platforms. Unlike traditional approaches that require reviewers to navigate between disparate systems, Octozi creates a unified data model that preserves source traceability while enabling cross-domain analysis.

\subsubsection{Contextual Intelligence Engine}
A distinguishing feature of the platform is its ability to automatically contextualize each data point within the patient's clinical journey. The system employs temporal reasoning algorithms to identify clinically relevant relationships between events, such as:
\begin{itemize}
    \item Temporal associations between laboratory abnormalities and medication changes (e.g., elevated liver enzymes occurring 72 hours after initiating a hepatotoxic medication)
    \item Pattern recognition across multiple parameters (e.g., correlating weight gain, edema reports, and elevated BNP levels to identify potential heart failure)
\end{itemize}

This contextual analysis extends beyond simple temporal proximity to incorporate clinical significance scoring based on established medical guidelines and expert-encoded rules. For example, an abnormal hemoglobin level in a patient with a history of anemia is identified with a different clinical significance than a patient experiencing a sudden drop in hemoglobin.

\subsubsection{Intelligent Discrepancy Detection}
Octozi employs proprietary large language models (LLMs) derived from Llama 4, specifically fine-tuned for clinical trial data cleaning tasks \cite{metaLlamaHerd}. The discrepancy detection engine combines these LLMs with a suite of heuristic algorithms, systematically developed in collaboration with expert medical data reviewers. The hybrid approach enables the platform to efficiently identify complex data inconsistencies by integrating domain-specific clinical reasoning with advanced language model capabilities. When evaluated on the annotated synthetic dataset, the discrepancy detection algorithms achieved a classification accuracy of 83.6\%, with a recall of 97.5\% and precision of 77.2\%. Importantly, the accuracy metric reflects the correct prediction of the specific category of discrepancies, rather than the simple detection of anomalies, underscoring the model's ability to replicate nuanced clinical decision-making patterns.

\subsubsection{Interactive Review Interface}
The user interface presents identified discrepancies and contextual information through multiple complementary views. A tabular view provides structured access to raw data with color-coded severity indicators and filtering capabilities. Moreover, a narrative view generates human-readable summaries that explain the clinical context and rationale for the flagged items, mimicking the mental model of experienced reviewers. This dual presentation format accommodates different cognitive styles and review scenarios.

The platform also incorporates query generation assistance, providing suggested query text based on the specific discrepancy type and regulatory requirements. This feature standardizes query language across reviewers while reducing the time required for query composition. Reviewers retain full control to modify or reject suggested queries, maintaining human oversight of all decisions.

Visual examples of the platform's interface and workflow are provided in Appendix B, demonstrating how the system presents complex clinical relationships in an intuitive, actionable format that accelerates decision-making while maintaining audit trails for regulatory compliance.

\section{Results}\label{sec3}
\subsection{Data Cleaning Efficiency}
Analysis of individual participant performance revealed consistent and substantial improvements across all reviewers (Fig. \ref{fig:participant_improvement}). Using traditional spreadsheet methods, participants correctly cleaned a mean of 3.4 data points (range: 1-6), while with Octozi assistance, this increased to a median of 20.5 data points (range: 7-35) within the same 30-minute review period. The improvement ratios ranged from 3.7x to 15x, with a mean improvement of 6.03x across all participants.

\begin{figure}
    \centering
    \includegraphics[width=1.0\linewidth]{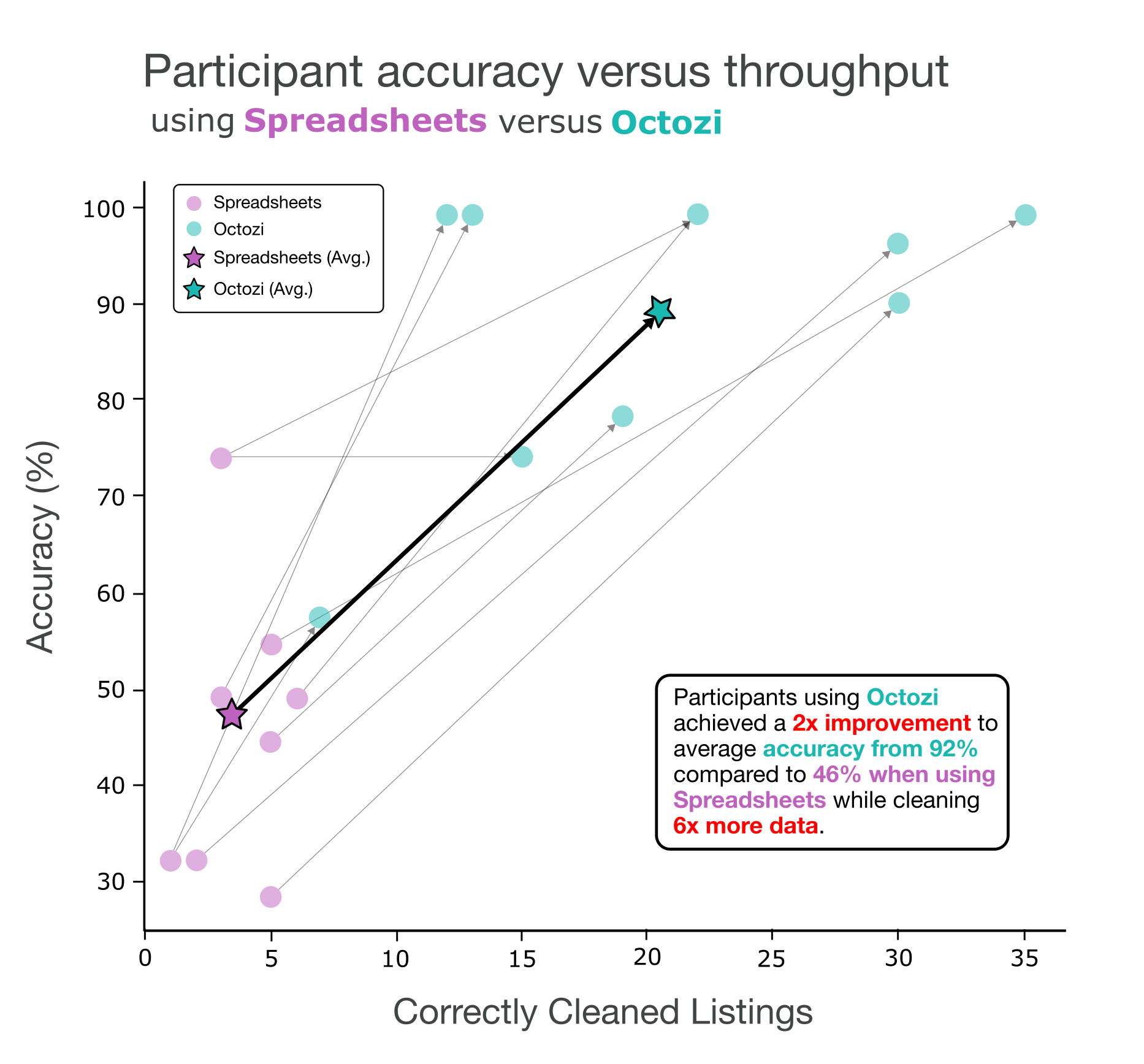}
    \caption{Accuracy versus throughput for individual participants using traditional spreadsheets and the Octozi AI-assisted platform. Each arrow represents the change in performance for a single participant, demonstrating consistent improvements in both accuracy and throughput with Octozi. On average, participants achieved 92\% accuracy while cleaning six times more data compared to 46\% accuracy with spreadsheets. Notably, no participant experienced a trade-off between speed and accuracy; all participants improved on both dimensions, with the lowest Octozi performance exceeding the highest spreadsheet performance.}
    \label{fig:accuracy_throughput_balance}
\end{figure}

Notably, the visualization of participant trajectories (Fig. \ref{fig:accuracy_throughput_balance}) demonstrates that every single participant experienced both improved accuracy and increased throughput when using Octozi, with no observed traded-off between speed and accuracy. The consistent upward and rightward movement of all participant data points illustrates the dual benefit of AI assistance, faster and more accurate data cleaning. The worst performer using Octozi performed better than the best participant using spreadsheets, highlighting the substantial and consistent improvement.

\subsection{Reduction in Medical Review Error Rates}
\begin{figure*}
    \centering
    \includegraphics[width=1.0\linewidth]{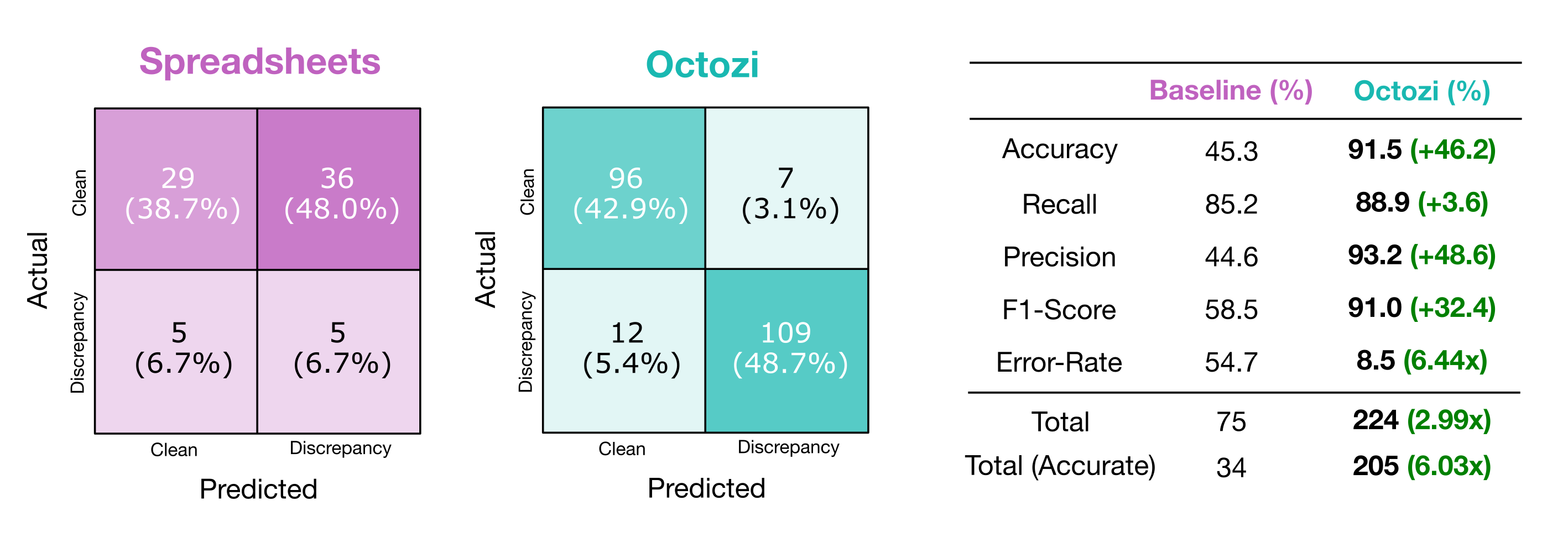}
    \caption{Confusion matrices and performance metrics comparing manual spreadsheet review and AI-assisted review using Octozi. Octozi substantially improved classification performance, increasing overall accuracy from 45.3\% to 91.5\% (+46.2 percentage points) and precision from 44.6\% to 93.2\% (+48.6 percentage points), while maintaining high recall. The F1-score improved from 58.5 to 91.0, reflecting balanced gains in precision and recall. Notably, false positive classifications—clean data incorrectly flagged as discrepancies—were reduced 15-fold (48.0\% to 3.1\%), indicating a significant reduction in erroneous queries and potential site burden.}
    \label{fig:CM}
\end{figure*}

The confusion matrices reveal a transformative improvement in clinical data cleaning accuracy through AI assistance (Fig. \ref{fig:CM}). Most notably, the error rate plummeted from 54.7\% with manual review to just 8.5\% with Octozi assistance, representing a 6.44-fold reduction in errors. This dramatic improvement means that reviewers using AI assistance made mistakes on fewer than 1 in 11 data points, compared to errors on more than half of all data points when using traditional spreadsheet methods.

This error reduction was driven by substantial improvements across all performance metrics. Overall accuracy improved from 45.3\% to 91.5\%, representing a 46.2 percentage point absolute improvement. The enhancement was particularly pronounced in precision, which increased from 44.6\% to 93.2\% (48.6 percentage point improvement), indicating that AI assistance nearly eliminated false positive identifications. When reviewers flagged a discrepancy using Octozi, they were correct 93\% of the time, compared to being wrong more often than right (44.6\% precision) with manual methods.

The F1-score, which balances precision and recall, improved from 58.5 to 91.0 (32.4 point increase), demonstrating that AI assistance enhanced both the completeness and accuracy of discrepancy detection. Interestingly, recall remained high in both conditions (85.2\% vs 88.9\%), suggesting that experienced reviewers were already adept at recognizing potential issues, but AI assistance helped them correctly distinguish true discrepancies from false alarms. The 6.44-fold reduction in error rate represents not just an incremental improvement but a fundamental shift in review reliability, transforming clinical data cleaning from an error-prone process to a highly accurate operation.

\subsection{Reduction in False Positive Queries: Minimizing Site Burden}
Perhaps most significantly for operational efficiency, the confusion matrices demonstrate a dramatic reduction in false positive classifications. With spreadsheet-based review, 36 out of 75 total classifications (48.0\%) were false positives—instances where reviewers incorrectly flagged clean data as discrepant. With Octozi, this dropped to just 7 out of 224 classifications (3.1\%), representing a 15.48-fold reduction in false query generation.

This reduction has profound implications for trial operations. In the manual condition, nearly half of all queries sent to sites would have been unnecessary, creating substantial bottlenecks to overburdened sites. The Octozi platforms's ability to maintain high sensitivity while dramatically improving specificity addresses a fundamental challenge in clinical data management.

\subsection{Statistical Significance and Effect Sizes}
All primary performance differences between methods achieved statistical significance with large effect sizes. The accuracy improvement yielded a Cohen's d of 2.8, indicating an extremely large effect size that represents meaningful practical significance beyond statistical significance. Similarly, throughput improvements demonstrated effect sizes exceeding 3.0, confirming that observed differences reflect genuine performance advantages rather than statistical artifacts.
Paired t-test analysis confirmed that improvements were consistent across participants (p < 0.001 for all primary measures), with 95\% confidence intervals that excluded the possibility of equivalent performance between methods. The magnitude and consistency of observed improvements provide strong evidence for the superiority of AI-assisted medical data review in both efficiency and accuracy domains.

\begin{figure}
    \centering
    \includegraphics[width=1.0\linewidth]{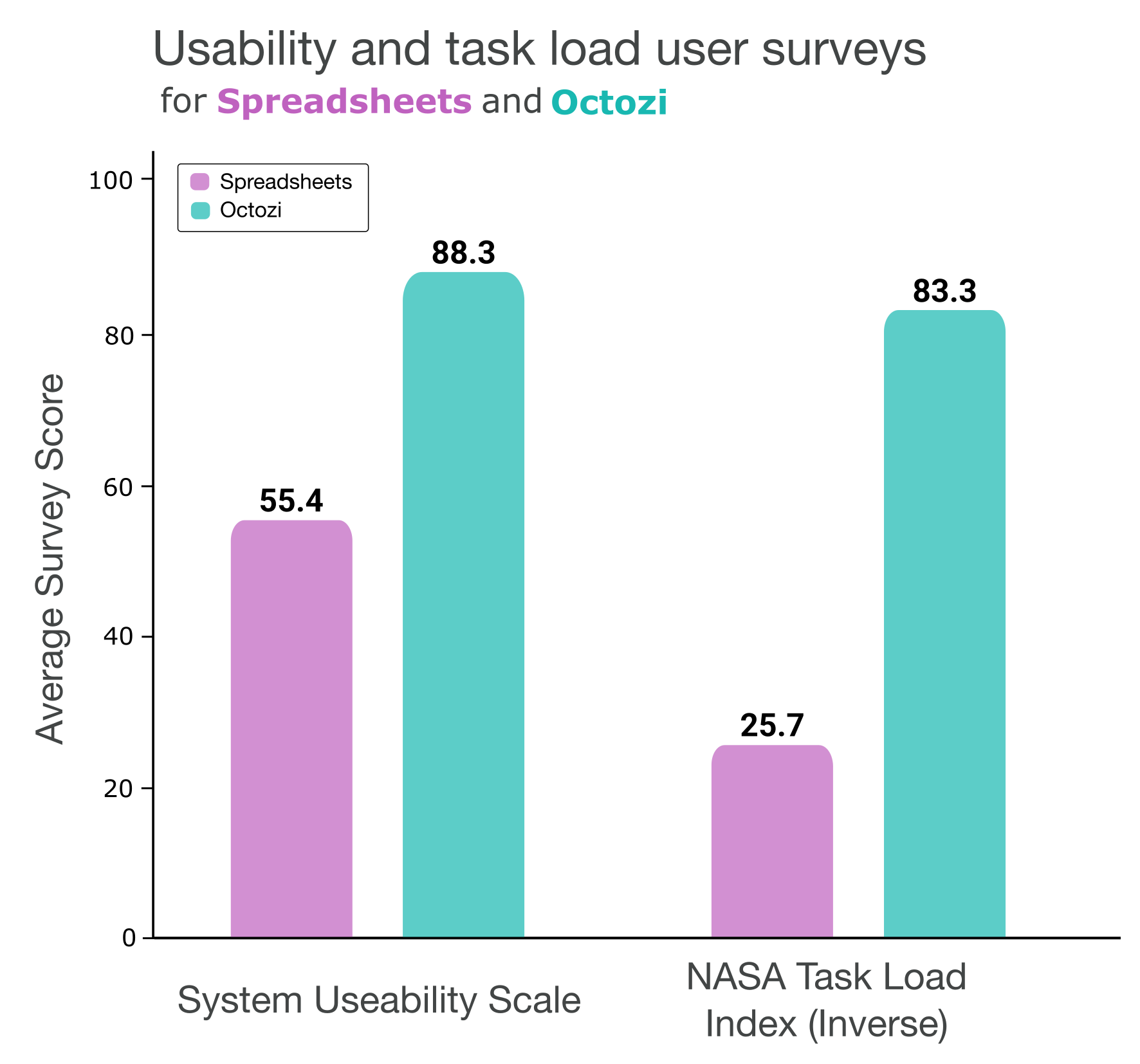}
    \caption{System Usability Scale (SUS) scores demonstrate superior usability of Octozi (88.3) compared to traditional spreadsheets (55.4), with scores above 68 considered acceptable and above 80 considered excellent. Inverse NASA Task Load Index (NASA-TLX) scores reveal a 3.2-fold reduction in cognitive workload with Octozi (83.3) versus spreadsheets (25.7), where higher scores indicate less mental demand, physical demand, temporal demand, performance concerns, effort, and frustration.}
    \label{fig:participant_surveys}
\end{figure}

\subsection{Usability and User Experience}
Post-study usability assessments revealed unanimous preference for the AI-assisted system across all participants. The inverse NASA Task Load Index (NASA-TLX) demonstrated a dramatic reduction in cognitive workload, with Octozi achieving a mean score of 83.3 compared to 25.7 for traditional spreadsheet methods (higher scores indicate reduced burden). This 3.2-fold improvement in task load represents a fundamental shift from high cognitive demand to manageable workload, with participants reporting substantial improvements across all six NASA-TLX dimensions: mental demand, physical demand, temporal demand, performance concerns, effort, and frustration.

System Usability Scale (SUS) assessments further validated the superior user experience of the AI-assisted approach. Octozi achieved a mean SUS score of 88.3, placing it in the "A+" category (scores above 84.1), while traditional spreadsheet methods scored 55.4, falling into the "D" category (scores between 51.7-62.6) \cite{lewis2018item}. This 32.9-point difference represents not merely an incremental improvement but a transformation from a "D" to an "A+" user experience (Fig. \ref{fig:participant_surveys}).

Qualitative feedback consistently highlighted three key advantages of the AI system. First, participants valued the intelligent prioritization of findings, which directed attention to the most clinically significant discrepancies rather than requiring sequential review of all data points. Second, the provision of contextual clinical information eliminated the need for constant reference switching between multiple data sources, reducing both time and cognitive load. Third, the system's ability to surface complex cross-domain discrepancies revealed patterns that participants acknowledged would be "nearly impossible to detect through manual review in the same time," as noted by one senior medical monitor with 6 years of experience.

\section{Discussion}\label{sec4}
Our findings demonstrate that AI assistance can fundamentally transform medical data review from a rate-limiting bottleneck into an accelerator of drug development. The magnitude of improvement observed over 6-fold increase in throughput coupled with doubling of accuracy, exceeds typical incremental advances in clinical trial operations and suggests a true paradigm shift in how data quality assurance can be conducted.

\subsection{Implications for Clinical Trial Operations}
The demonstrated improvements have immediate and far-reaching implications for clinical trial conduct. The 6-fold improvement in data review throughput could compress critical trial milestones that currently delay drug development by weeks or months. Database preparation for interim analyses, which traditionally requires 4-8 weeks of intensive effort, could potentially be completed within 1-2 weeks using AI-assisted methods. This acceleration is particularly crucial for adaptive trial designs that rely on timely interim data to modify study parameters, and for rare disease trials where patient populations face urgent medical needs.

The 46.2\% improvement in accuracy, combined with the 15.48-fold reduction in false positive queries, addresses two fundamental challenges in clinical data management: maintaining high data quality standards while minimizing site burden. Current industry practice shows that medical reviewers generate queries with only 51\% accuracy \cite{stokman2021risk}, creating substantial administrative burden for clinical sites. Our results demonstrate that AI assistance can nearly eliminate unnecessary queries while maintaining high sensitivity for true discrepancies, potentially transforming the relationship between sponsors and investigative sites.

\begin{figure}
    \centering
    \includegraphics[width=1.0\linewidth]{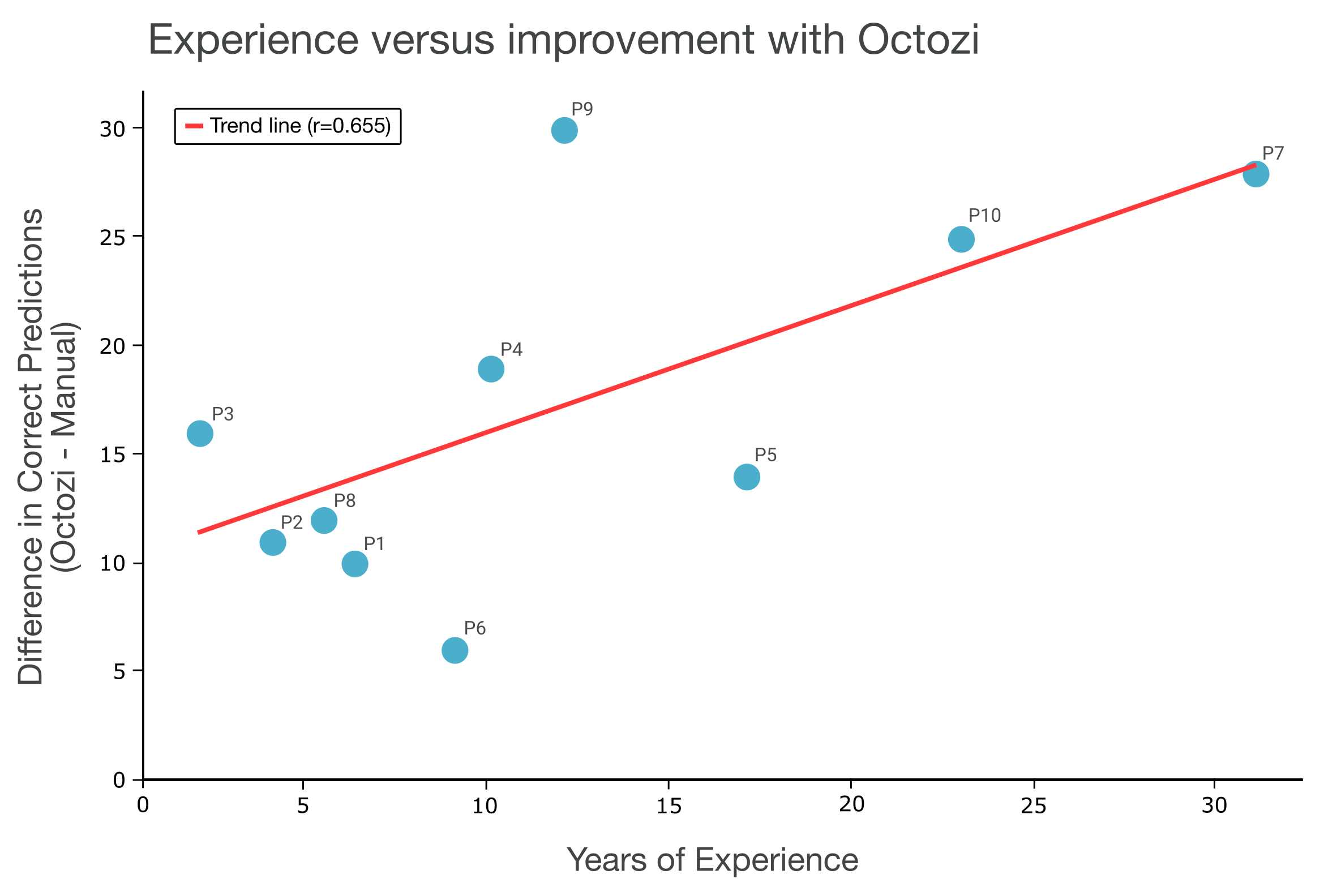}
    \caption{Relationship between reviewer experience and performance improvement with AI augmentation. The positive trend line (r=0.655) indicates that more experienced reviewers achieved greater absolute improvements when using AI assistance. This pattern suggests that AI augmentation serves as a force multiplier for clinical expertise rather than merely a substitute for experience.}
    \label{fig:experience_improvement}
\end{figure}

\subsection{Economic Impact Analysis}
To quantify the economic implications of our findings, we conducted a comprehensive cost-benefit analysis for a representative Phase III oncology trial with 1,100 patients generating approximately 4.5 million data points over four years. Table \ref{tab:economic_impact} summarizes the primary areas of cost savings achieved through AI-assisted data review. The analysis reveals three primary areas of economic impact:

\begin{table*}[]
    \caption{Economic impact analysis of AI-assisted medical data cleaning for a representative Phase III oncology trial (1,100 patients, 4.5M data points, 4-year duration). Conservative estimates assume 50\% realization of observed experimental improvements to account for real-world implementation factors.}
    \label{tab:economic_impact}
    \begin{tabular}{@{}lccccc@{}}
        \toprule
        Cost Category & Traditional Cost & AI-Assisted Cost & Savings & \% Reduction & Basis of Calculation \\
        \midrule
        \textbf{Medical Review} & & & & & \\
        Review hours & 2,100 hrs & 700 hrs & 1,400 hrs & 66.7\% & 3x efficiency gain \\
        Total cost (\$300/hr) & \$630,000 & \$210,000 & \$420,000 & 66.7\% & (conservative vs. 6.03x observed) \\
        \midrule
        \textbf{Query Management} & & & & & \\
        Total queries generated & 5,154 & 2,881 & 2,273 & 44.1\% & 10x false positive reduction \\
        False positive queries & 2,526 & 253 & 2,273 & 90.0\% & (conservative vs. 15.48x observed) \\
        Total cost (\$126.50/query) & \$651,949 & \$364,439 & \$287,510 & 44.1\% & 56\% reduction in queries \\
        \midrule
        \textbf{Database Lock} & & & & & \\
        Timeline (days) & 36.8 & 31.8 & 5.0 & 13.6\% & Conservative estimate \\
        Lost revenue (\$840K/day) & \$30,912,000 & \$26,712,000 & \$4,200,000 & 13.6\% & (expert panel: 5-12 day range) \\
        Operational cost (\$40K/day) & \$1,472,000 & \$1,272,000 & \$200,000 & 13.6\% & \\
        Subtotal & \$32,384,000 & \$27,984,000 & \$4,400,000 & 13.6\% & \\
        \midrule
        \textbf{Total Direct Savings} & \$33,665,949 & \$28,558,439 & \textbf{\$5,107,510} & \textbf{15.2\%} & \\
        \bottomrule
    \end{tabular}
\end{table*}

\textbf{Medical Review Efficiency}: Traditional medical review requires approximately 2,100 hours across three reviewers, costing \$630,000. Applying a conservative 3-fold efficiency improvement (half of our observed 6.03-fold gain), total review time reduces to 700 hours, yielding \$420,000 in savings.

\textbf{Query Management Optimization}: With medical reviewers generating and managing 5,154 queries at \$126.50 each, traditional approaches cost \$651,949. Our demonstrated 15.48-fold reduction in false positives (conservatively adjusted to 10-fold) reduces total queries by 56\%, saving \$287,510 while maintaining high sensitivity for true discrepancies.

\textbf{Database Lock Acceleration}: The most substantial savings derive from accelerated database lock. A conservative 5-day reduction from the typical 36.8-day timeline saves \$4.4 million through earlier market entry (\$840,000/day in opportunity cost) and reduced operational expenses (\$40,000/day).

The cumulative direct savings of \$5.1 million represent only quantifiable operational improvements. Additional unquantified benefits include improved regulatory submission quality, earlier safety signal detection, reduced protocol deviations, and enhanced site relationships through reduced query burden. A detailed economic analysis is provided in Appendix A.

\subsection{Human-AI Collaboration Model}
Our results provide compelling evidence for a collaborative model where AI augments rather than replaces human expertise. The positive correlation between reviewer experience and AI-enabled performance gains (r=0.655, Fig. \ref{fig:experience_improvement}) suggests that AI serves as a force multiplier for clinical expertise. Experienced reviewers achieved the greatest absolute improvements, gaining up to 28 additional correct predictions compared to 12 for novice reviewers, indicating that domain knowledge remains crucial for maximizing AI benefits.

This collaborative model addresses key concerns about AI adoption in regulated environments. By maintaining human oversight while automating routine tasks, the system preserves the clinical judgment essential for complex medical decisions while eliminating the mechanical aspects of data review. The educational benefits observed where participants identified previously unrecognized discrepancy types further suggest that AI systems can enhance professional development while improving operational efficiency.



\section{Conclusion}\label{sec5}

The integration of AI into clinical trial data cleaning represents a transformative opportunity to address fundamental inefficiencies in drug development. Our controlled experimental study provides robust evidence that AI assistance can simultaneously improve the speed, accuracy, and consistency of medical data review while reducing operational burden on clinical trial sites. The observed 6.03-fold improvement in throughput, coupled with a doubling of accuracy and 15.48-fold reduction in false positive queries, demonstrates the potential for AI to accelerate drug development timelines by weeks to months while maintaining or improving data quality standards.

These findings offer immediate practical implications for the pharmaceutical industry, with potential cost savings exceeding \$5 million per Phase III trial through improved efficiency, reduced query burden, and accelerated database lock. More broadly, our results demonstrate the potential for human-AI collaboration to enhance performance in complex, safety-critical domains requiring both efficiency and expert judgment.

As the pharmaceutical industry grapples with increasing trial complexity and data volumes, AI-assisted approaches offer a scalable solution that enhances rather than replaces human expertise. The consistent benefits observed across reviewer experience levels suggest that AI can democratize access to high-quality data review capabilities, potentially enabling more efficient global clinical trials and ultimately accelerating the delivery of life-saving therapies to patients in need.

\bibliographystyle{ACM-Reference-Format}
\bibliography{sample-base}

\newpage\phantom{blabla}
\appendix

\section{Detailed Economic Impact Analysis}

\subsection{Methodology and Assumptions}
This economic analysis evaluates the financial implications of implementing AI-assisted medical data cleaning in a representative Phase III oncology trial. Our model is based on:

\begin{itemize}
    \item Trial parameters: 1,100 patients, 4.5 million data points, 4-year duration
    \item Industry benchmarks from published literature and proprietary databases
    \item Expert panel consisting of three senior medical monitors (average 20+ years experience)
    \item Conservative adjustment factors applied to experimental results
\end{itemize}

\subsection{Detailed Cost Breakdown}

\subsubsection{Medical Review Labor Costs}
Traditional medical review requirements were calculated based on insight from the expert panel:

\begin{itemize}
    \item \textbf{Baseline workload}: 150-200 hours per reviewer per year (used 175)
    \item \textbf{Team size}: 3 medical reviewers (standard for oncology Phase III)
    \item \textbf{Total hours}: 2,100 hours over 4 years
    \item \textbf{Hourly rate}: \$300 (rate for an experienced senior medical monitor at a sponsor)
    \item \textbf{Total traditional cost}: \$630,000
\end{itemize}

AI-assisted projections assume a conservative 3-fold efficiency gain (50\% of the 6.03-fold improvement observed experimentally), accounting for:
\begin{itemize}
    \item Data cleaning and review is typically done in batches
    \item Rounding up data review hours over the course of trial
    \item Estimated reviewer fatigue for long session over 1 hours
\end{itemize}

\subsubsection{Query Management Economics}
Query generation and resolution represents a significant cost center:

\begin{itemize}
    \item \textbf{Total data points}: 4,341,900 (based on 4100 fields per patient and 1100 patients in this study)
    \item \textbf{Manual query rate}: 20\% of queries are estimated to be manual as opposed to automated edit checks \cite{lewis2018item}
    \item \textbf{Industry query rate}: 7.3\% of data points for oncology study \cite{stokman2021risk}
    \item \textbf{Medical reviewer queries}: 8.13\% of total queries \cite{stokman2021risk}
    \item \textbf{Cost per query}: \$126.50 (average of \$28-\$225 range) \cite{stokman2021risk}
\end{itemize}

Query cost components include:
\begin{itemize}
    \item Initial query generation and documentation
    \item Clinical team review and approval
    \item Site communication and follow-up
    \item Response review and possible escalation
\end{itemize}

The 15.48-fold reduction in false positives observed experimentally was conservatively adjusted to 10-fold for real-world projections, considering:
\begin{itemize}
    \item Medical reviewers aiming to clarify information instead of disputing data
    \item Estimated efficiency losses
\end{itemize}

The savings related to query management for medical reviewers considers:
\begin{itemize}
    \item \textbf{\% of no-change queries} 51\% of queries created by medical reviewers do not result in changes to the dataset \cite{stokman2021risk}
    \item \textbf{Number of false queries} 2,525 ($5,154 * 0.51$)
    \item \textbf{Reduction in false query rate} 10-fold decrease (conservative estimate)
    \item \textbf{Remaining false queries} 253 ($2,525 / 10$)
\end{itemize}

\subsubsection{Database Lock Value Analysis}
Database lock acceleration provides the highest economic value:

\begin{itemize}
    \item \textbf{Industry average DBL}: 36.8 days (range: 2-120 days) \cite{harper2021characterizing}
    \item \textbf{Conservative reduction}: 5 days (expert panel estimated 5-12 days)
    \item \textbf{Daily opportunity cost}: \$840,000 (median loss of revenue from delayed trial cost for oncology trial \cite{smith2024new})
    \item \textbf{Daily operational cost}: \$40,000 (clinical team costs, manufacturing, sites, etc. \cite{smith2024new})
\end{itemize}

\subsection{Indirect Benefits Not Quantified}
Several additional benefits were not included in the financial analysis:

\begin{itemize}
    \item \textbf{Improved safety signal detection}: Earlier identification of adverse events could prevent costly late-stage failures (average Phase III failure costs \$150M)
    \item \textbf{Enhanced regulatory submission quality}: Reduced likelihood of Complete Response Letters (CRL), each potentially costing \$50-100M in delays
    \item \textbf{Site relationship improvements}: Reduced query burden may improve site retention and recruitment for future trials
    \item \textbf{Competitive advantage}: Faster trial completion enables earlier market entry and extended market exclusivity
    \item \textbf{Portfolio effects}: Efficiency gains enable reallocation of resources to additional pipeline programs
\end{itemize}

Total implementation investment of approximately \$400,000 is recovered within the first month of database lock acceleration.

\section{Enrollment Questionnaire}
Table \ref{tab:questionnaire} describes the question asked to interested participants in the study and the type of response for that question. These questions help ascertain if a potential participant was eligible for the study, contact information, and information to confirm their background. 

\begin{table*}[]
\caption{Screening questions asked to respondents.}
\begin{tabular}{|l|l|}
\hline
Question & Field Type \\ \hline
First and last name & Text \\ \hline
Email & Text \\ \hline
LinkedIn & Text \\ \hline
Resume / CV & File \\ \hline
How would you describe your current or most recent role? (Select all that apply) & Checkboxes \\ \hline
Does your current or most recent role involve medical cleaning or reviewing of clinical trial data? & Yes/No \\ \hline
How many years have you worked in roles involving data cleaning or clinical data review? & \textless{}2, 2-5, 5+ years \\ \hline
Roughly how many clinical trials have you been involved in where you were responsible for medical data cleaning? & \textless{}2, 2-5, 5+ years \\ \hline
Have you had experience with medical data cleaning/review for oncology studies? & Yes/No \\ \hline
Which tools have you used for data cleaning or review? (Select all that apply) & Checkboxes \\ \hline
Why are you interested in participating in this study? & Text \\ \hline
\end{tabular}
\label{tab:questionnaire}
\end{table*}

\section{Octozi Platform}
\begin{figure*}[h!]
    \centering
    \includegraphics[width=0.95\linewidth]{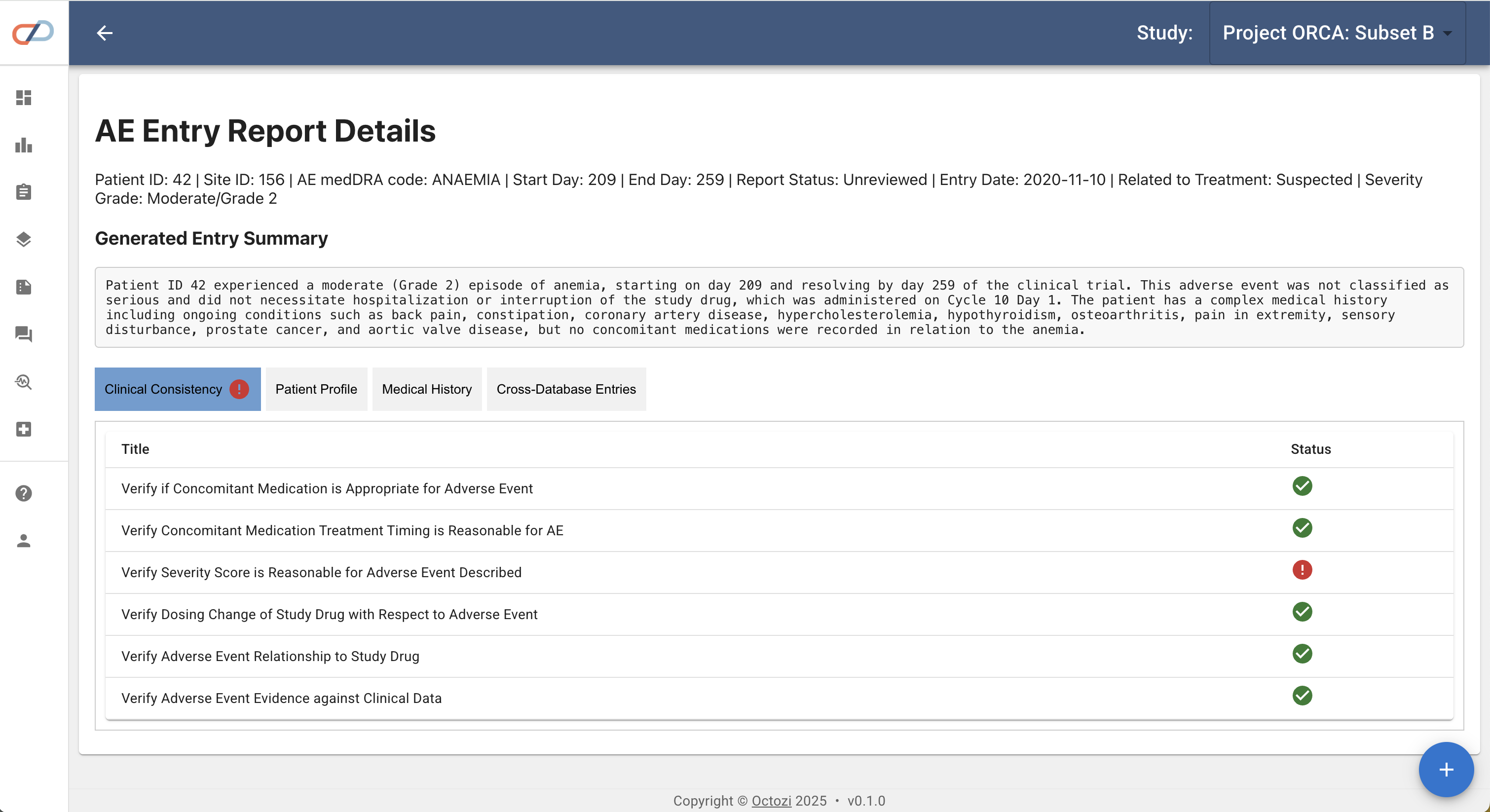}
    \caption{A screenshot of the Octozi platform for an adverse event. The page presents the user with a short narrative describing the context for the adverse event of anemia, discrepancies found by the platform, and all of the patient information as figures in "Patient Profile" or in tabular form in "Cross-Database Entries". }
    \label{fig:octozi_platform_1}
\end{figure*}

\begin{figure*}[h!]
    \centering
    \includegraphics[width=0.95\linewidth]{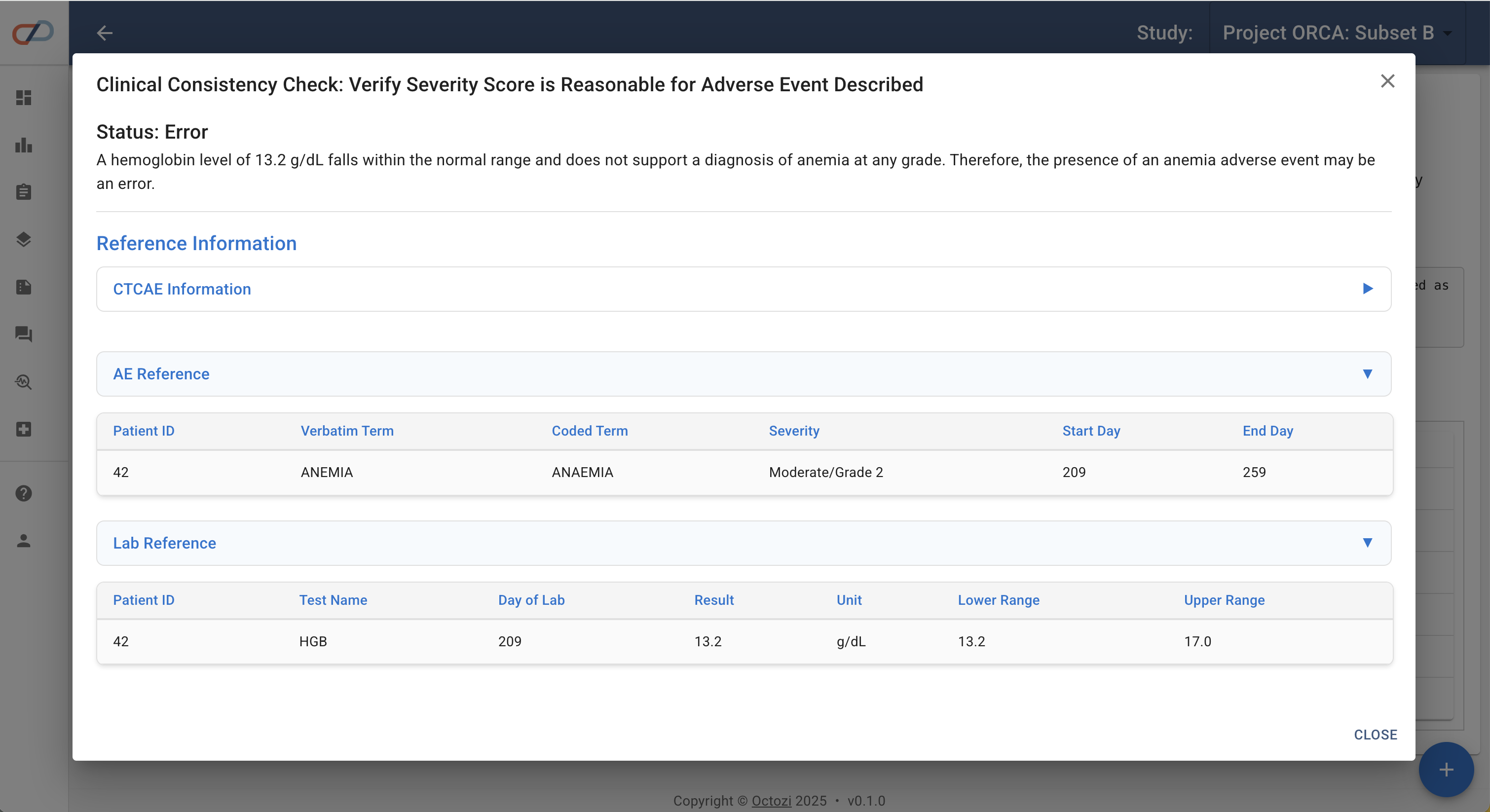}
    \caption{A screenshot of the Octozi platform for a discrepancy check that has found a potential inconsistency in the data. The context engine has pulled relevant adverse event and laboratory case report forms with limited column values for the adverse event in question. Not shown are the CTCAE reference ranges and text explanations for anemia. The output of the AI-assistant is provided at the top of the reference information to direct the user to its conclusion.}
    \label{fig:octozi_platform_2}
\end{figure*}

Figures \ref{fig:octozi_platform_1} and \ref{fig:octozi_platform_2} are screenshots of the Octozi platform that a user would see after selecting a data entry (one case report entry) in a previous menu. What is not presented in these figures are sorting algorithms that present the user with critical entries that contain potential discrepancies first which lower priority and likely clean data last. 

\end{document}